\documentclass[10pt,twocolumn,letterpaper]{article}

\usepackage{iccv}
\usepackage{times}
\usepackage{epsfig}
\usepackage{graphicx}
\usepackage{amsmath}
\usepackage{amssymb}

\usepackage{color,xcolor}
\usepackage{multirow}
 
\usepackage{makecell}

\usepackage{algorithm}
\usepackage{algorithmic}

\usepackage{bm}
\usepackage{dsfont}
\usepackage{booktabs}

\usepackage{color}
\usepackage{xcolor}

\usepackage[pagebackref=true,breaklinks=true,letterpaper=true,colorlinks,bookmarks=false]{hyperref}

\iccvfinalcopy 

\def\iccvPaperID{3743} 

\ificcvfinal\fi

\begin{document}

\title{RepQ-ViT: Scale Reparameterization for Post-Training Quantization of \\ Vision Transformers}

\author{Zhikai Li$^{1,2}$, Junrui Xiao$^{1,2}$, Lianwei Yang$^{1,2}$, and Qingyi Gu$^{1,}$\thanks{Corresponding author.} \\
$^1$Institute of Automation, Chinese Academy of Sciences\\
$^2$School of Artificial Intelligence, University of Chinese Academy of Sciences\\
{\tt\small \{lizhikai2020, xiaojunrui2020, yanglianwei2021, qingyi.gu\}@ia.ac.cn}}

\maketitle
\ificcvfinal\thispagestyle{empty}\fi

\begin{abstract}
  Post-training quantization (PTQ), which only requires a tiny dataset for calibration without end-to-end retraining, is a light and practical model compression technique. Recently, several PTQ schemes for vision transformers (ViTs) have been presented; unfortunately, they typically suffer from non-trivial accuracy degradation, especially in low-bit cases. In this paper, we propose RepQ-ViT, a novel PTQ framework for ViTs based on quantization scale reparameterization, to address the above issues. RepQ-ViT decouples the quantization and inference processes, where the former employs complex quantizers and the latter employs scale-reparameterized simplified quantizers. This ensures both accurate quantization and efficient inference, which distinguishes it from existing approaches that sacrifice quantization performance to meet the target hardware. More specifically, we focus on two components with extreme distributions: post-LayerNorm activations with severe inter-channel variation and post-Softmax activations with power-law features, and initially apply channel-wise quantization and log$\sqrt{2}$ quantization, respectively. Then, we reparameterize the scales to hardware-friendly layer-wise quantization and log2 quantization for inference, with only slight accuracy or computational costs. Extensive experiments are conducted on multiple vision tasks with different model variants, proving that RepQ-ViT, without hyperparameters and expensive reconstruction procedures, can outperform existing strong baselines and encouragingly improve the accuracy of 4-bit PTQ of ViTs to a usable level. Code is available at \url{https://github.com/zkkli/RepQ-ViT}.
\end{abstract}

\section{Introduction}
With the powerful representational capabilities of the self-attention mechanism, vision transformers (ViTs) have recently demonstrated surprising potential in a range of vision applications, including image classification \cite{dosovitskiy2020image,liu2021swin}, object detection \cite{carion2020end,zhu2020deformable}, semantic segmentation \cite{strudel2021segmenter}, etc., and are thus being widely investigated as new vision backbones \cite{han2022survey}.
However, ViTs rely on heavy and intensive computations, resulting in intolerable memory footprint, power consumption, and inference latency, which hinders their deployment on resource-constrained edge devices \cite{hou2022multi,tang2022patch}. Consequently, compression techniques for ViTs are essential in real-world applications, particularly where low-cost deployment and real-time inference are desired.

Model quantization, which reduces model complexity by decreasing the representation precision of weights and activations, is an effective and prevalent compression approach \cite{gholami2021survey,krishnamoorthi2018quantizing}. A notable research line is based on quantization-aware training (QAT) \cite{choi2018pact,esser2019learned}, which relies on end-to-end retraining to compensate for the accuracy of the quantized model. Despite the good performance, such retraining requires gradient backpropagation and parameter updates on the entire training dataset, which brings undesirably large time and resource costs \cite{li2022dual,wei2022qdrop}. 
Fortunately, another family of methods, referred to as post-training quantization (PTQ), can overcome the above challenges \cite{wang2020towards,li2021brecq,nagel2020up}. It simply takes a tiny unlabeled dataset to calibrate the quantization parameters without retraining and thus is regarded as a promising and practical solution. 

Although various PTQ methods for convolutional neural networks (CNNs) have been proposed in previous works with good performance, they produce disappointing results on ViTs, with more than 1\% accuracy drop even in 8-bit quantization \cite{yuan2021ptq4vit}. To this end, several efforts identify the key components that limit the quantization performance of ViTs, such as LayerNorm, Softmax, and GELU, and propose PTQ schemes accordingly \cite{lin2021fq,liu2021post}.
Nevertheless, when performing ultra-low-bit ($e$.$g$., 4-bit) quantization, the performance of these schemes is still far from satisfactory \cite{ding2022towards}.
The core reason for their low performance is that they invariably follow the traditional quantization paradigm, in which the initial design of the quantizers must account for the future inference overhead.
This forces previous methods to carefully design simple quantizers to accommodate the characteristics of the target hardware, even at the cost of remarkably sacrificing accuracy.

\emph{Is the traditional quantization-inference dependency paradigm the only option?} To answer this question, we explore the feasibility of decoupling the quantization and inference processes, and reveal that complex quantizers and hardware standards are not always antagonistic; instead, the two can be explicitly bridged via \emph{scale reparameterization}.
This potentially derives an interesting quantization-inference decoupling paradigm, in which complex quantizers are employed in the initial quantization to adequately preserve the original parameter distributions, and then they are transformed to simple hardware-friendly quantizers via scale reparameterization for actual inference, resulting in both high quantization accuracy and inference efficiency.

\begin{figure}[t]
    \centering
    \includegraphics[width=0.95\linewidth]{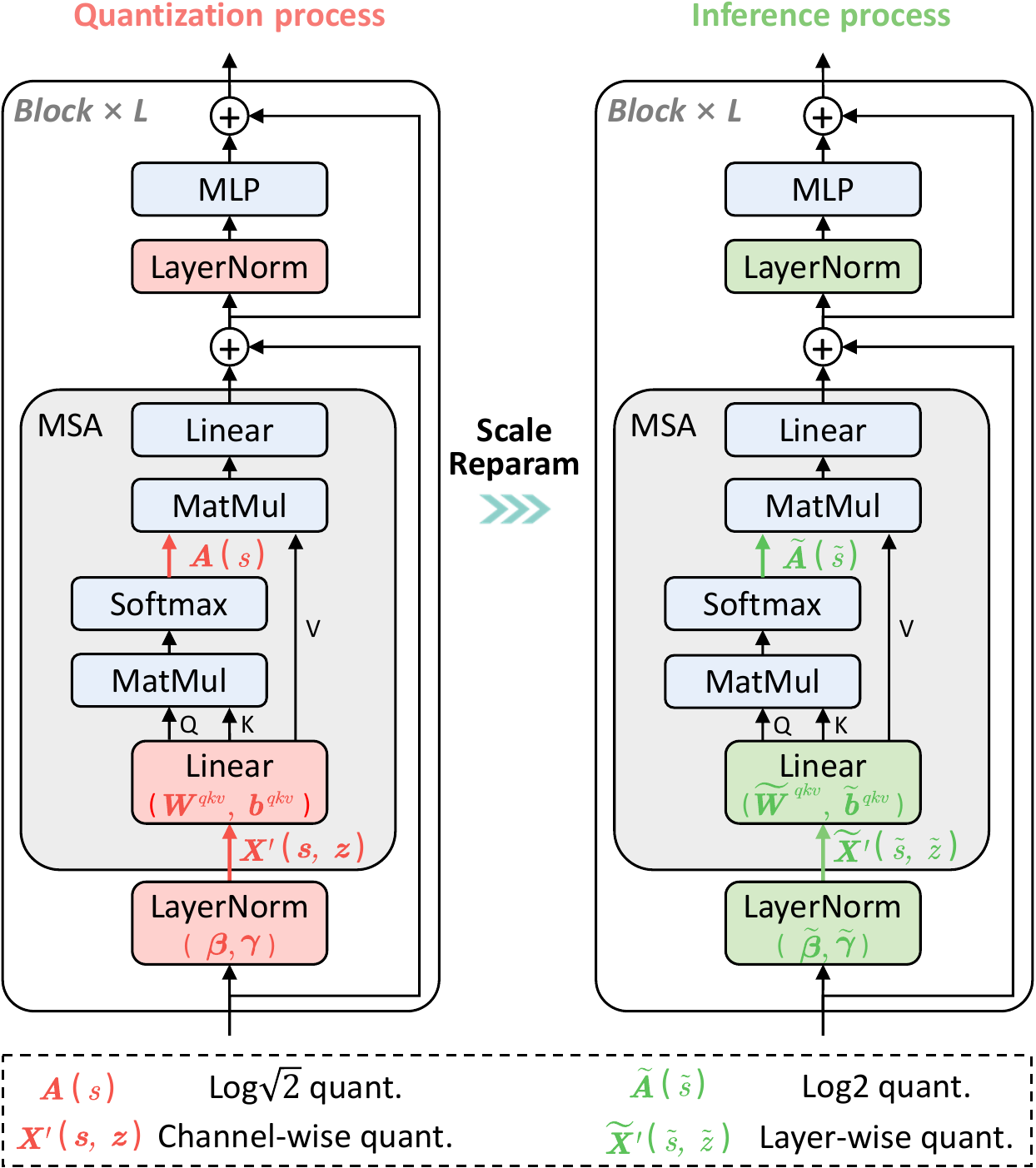}
    \caption{Overview of the RepQ-ViT framework. Building on the quantization-inference decoupling paradigm, for post-LayerNorm and post-Softmax activations, complex quantizers are employed in the quantization process and simple quantizers are employed in the inference process, with scale reparameterization bridging the two. }
\label{fig:overview}
\end{figure}

With the above insights, we propose a novel PTQ framework for ViTs, called RepQ-ViT, in this paper.
In RepQ-ViT, we focus on two components with extreme distributions in ViTs that challenge the direct use of simple quantizers.
Specifically, for post-LayerNorm activations, we initially apply channel-wise quantization to maintain their severe inter-channel variation, and then reparameterize the scales to layer-wise quantization to match the hardware, which is achieved by adjusting the LayerNorm's affine factors and the next layer's weights; for post-Softmax activations, since our study shows that their power-law distributions and the properties of attention scores prefer log$\sqrt{2}$ quantizers, we are interested in 
reparameterizing the scales to change the base to 2 to enable bit-shifting operations in inference. The overview of the RepQ-ViT framework is illustrated in Figure \ref{fig:overview}.
Note that the scale reparameterization methods presented in this paper enjoy theoretical support, with only a slight accuracy drop compared to complex quantizers or a slight computational overhead compared to simple quantizers, and thus have the potential to ensure interpretability and robustness. 

Our main contributions are summarized as follows:
\begin{itemize}
\item We propose a novel PTQ framework for ViTs that escapes from the traditional paradigm by decoupling the quantization and inference processes, with the former employing complex quantizers and the latter employing scale-reparameterized simplified quantizers, which has great potential in quantizing components with extreme distributions in ViTs.

\item For post-LayerNorm and post-Softmax activations, we initially apply channel-wise and log$\sqrt{2}$ quantization, respectively, to maintain the original data distribution, and then transform them to simple quantizers via interpretable scale reparameterization to match the hardware in inference. 

\item We evaluate RepQ-ViT on various vision tasks, including image classification, object detection, and instance segmentation, and RepQ-ViT, without hyperparameters and expensive reconstruction procedures, can encouragingly outperform existing baselines.
\end{itemize}

\section{Related Works}

\subsection{Vision Transformers}
ViTs, which exploit the self-attention mechanism to extract global information, have recently achieved excellent performance on a variety of computer vision tasks, showing great potential as general-purpose vision backbones \cite{han2022survey}.
ViT \cite{dosovitskiy2020image} attempts to remove all convolutions and apply a pure transformer-based model to the image classification task for the first time and achieves competitive results. Afterwards, several variants are proposed to further improve the performance. DeiT \cite{touvron2021training} introduces an efficient training strategy to reduce the dependency on large-scale training data, and Swin \cite{liu2021swin} applies a hierarchical architecture with shifted windows to enhance the modeling power of the self-attention mechanism.
In addition, ViTs have also been successfully applied to high-level applications, such as object detection \cite{carion2020end,zhu2020deformable} and semantic segmentation \cite{strudel2021segmenter}.

Despite the promising performance, the massive large matrix multiplications in ViTs incur huge memory footprints and computational overheads in real-world applications, which are intolerable in resource-constrained edge scenarios \cite{tang2022patch,hou2022multi}. Several works, such as MobileViT \cite{mehta2021mobilevit} and MiniViT \cite{zhang2022minivit}, attempt to address the above issues through lightweight architecture design, while they still provide substantial room for further compression as they keep floating-point parameters.

\subsection{Model Quantization}
Model quantization, which represents the original floating-point parameters with low-bit values, is an effective approach for compressing neural networks \cite{gholami2021survey,krishnamoorthi2018quantizing}.
To achieve competitive quantization performance, lots of methods follow the QAT pipeline and perform retraining on the entire training dataset \cite{choi2018pact,esser2019learned,wang2019haq,zhou2016dorefa}; however, such retraining is resource-intensive and time-consuming. Thus, PTQ, which is free from retraining, is believed to be a more promising solution for low-cost and rapid deployment.
Several impressive PTQ methods have been proposed with great success on CNNs, such as DFQ \cite{nagel2019data}, AdaRound \cite{nagel2020up}, and BRECQ \cite{li2021brecq}, yet they have poor performance on ViTs with substantially different structures.

As a result, designing PTQ methods for ViTs has recently received widespread interest.
Ranking loss \cite{liu2021post} is utilized to maintain the relative order of attention scores before and after quantization.
FQ-ViT \cite{lin2021fq} introduces Powers-of-Two Scale and Log-Int-Softmax to quantize LayerNorm and Softmax operations to obtain fully quantized ViTs.
PSAQ-ViT \cite{li2022patch,li2022psaq} designs a relative value metric to invert images and pushes PTQ for ViTs to data-free scenarios.
PTQ4ViT \cite{yuan2021ptq4vit} presents twin uniform quantization to cope with the unbalanced distributions of post-Softmax and post-GELU activations and uses a Hessian-guided metric to search for quantization scales. 
APQ-ViT \cite{ding2022towards} works on preserving the Matthew effect of post-Softmax activations and proposes a calibration scheme that perceives the overall quantization disturbance in a block-wise manner.
Unfortunately, the above methods produce non-trivial accuracy drops or even crashes in ultra-low-bit quantization. The main performance bottleneck stems from their direct use of simple hardware-oriented quantizers that cannot represent the extreme distributions well; in contrast, the novel paradigm proposed in this paper can potentially eliminate these issues.

\section{Methodology}

\paragraph{Overview}
Figure \ref{fig:overview} illustrates the overview of the proposed RepQ-ViT framework.
In the quantization-inference decoupling paradigm, the main challenge is to convert the initial complex quantizers to the simple quantizers for inference. Thus, we propose scale reparameterization methods for post-LayerNorm and post-Softmax activations, respectively, as detailed in Sections \ref{sec:layernorm} and \ref{sec:softmax}. Moreover, their flows are described in Algorithm \ref{alg:overview}.

\begin{algorithm}[t]\small
	\caption{Pipeline of RepQ-ViT framework.}
	\label{alg:overview}
	\begin{algorithmic}[1]
	   \STATE {\bfseries Input:} Pretrained full-precision model, Calib data

	   \STATE Initialize the quantized model with calib data and Eq. \ref{eq:sz}, where post-LayerNorm activations $\bm{X}'$ apply channel-wise quantization ($\bm{s}$, $\bm{z}$) and post-Softmax activations $\bm{A}$ apply log$\sqrt{2}$ quantization ($s$);

          \textcolor[RGB]{142,207,201}{\# Scale reparam for post-LayerNorm activations}

          \STATE Update the quantizer of $\bm{X}'$ via $\tilde{s}=\text{E}[\bm{s}]$ and $\tilde{z}=\text{E}[\bm{z}]$;
          
          \STATE Calculate $\bm{r}_1=\bm{s}/(\tilde{s}\cdot\bm{1})$ and $\bm{r}_2=\bm{z}-(\tilde{z}\cdot\bm{1})$;
          
          \STATE Update LayerNorm’s affine factors $\widetilde{\bm{\beta}}$ and $\widetilde{\bm{\gamma}}$ based on Eq. \ref{eq:raparm_1};

          \STATE Update next layer’s weights $\widetilde{\bm{W}}^{qkv}$ and $\widetilde{\bm{b}}^{qkv}$ based on Eq. \ref{eq:raparm_2};

          \STATE Re-calibrate $\widetilde{\bm{W}}^{qkv}$ with calib data;

          \textcolor[RGB]{142,207,201}{\# Scale reparam for post-Softmax activations}

          \STATE Update quantization procedure based on Eq. \ref{eq:3.3-1};

          \STATE Update $\tilde{s}$ in de-quantization procedure based on Eq. \ref{eq:3.3-2};
    
	   \STATE {\bfseries Output:} Quantized model
	\end{algorithmic}
\end{algorithm} 

\subsection{Preliminaries}
\paragraph{ViTs' standard structure} First, the input image is reshaped into $N$ flatted 2D patches, and they are subsequently projected by the embedding layer to a $D$-dimensional vector sequence, which is denoted as $\bm{X}_0\in \mathbb{R}^{N\times D}$ \footnote{To simplify the formulation, we ignore the batch dimension.}. Then, $\bm{X}_0$ is fed into a stack of transformer blocks, where each block consists of a multi-head self-attention (MSA) module and a multi-layer perceptron (MLP) module. With LayerNorm applied before each module and residuals added after each module, the transformer block is formulated as:
\begin{align}
  \bm{Y}_{l-1} & = \text{MSA}(\text{LayerNorm}(\bm{X}_{l-1})) + \bm{X}_{l-1} \\
  \bm{X}_l & = \text{MLP}(\text{LayerNorm}(\bm{Y}_{l-1})) + \bm{Y}_{l-1}
\end{align}
where $l=1,2,\cdots, L$, and $L$ is the number of the transformer blocks.

The MSA module learns inter-patch correlations of the input $\bm{X}'\in \mathbb{R}^{N\times D}$ through the following processes:
\begin{align}
    [\bm{Q}_i, \bm{K}_i, \bm{V}_i] &= \bm{X}'\bm{W}^{qkv}+\bm{b}^{qkv} \;\;\; i=1,2,\cdots,h\\
    \text{Attn}_{i} &= \text{Softmax}\left(\frac{\bm{Q}_i\cdot \bm{K}_i^T}{\sqrt{D_h}}\right)\bm{V}_i \\
    \text{MSA}(\bm{X}') & = [\text{Attn}_{1},\text{Attn}_{2},\ldots, \text{Attn}_{h}] \bm{W}^o + \bm{b}^o
\end{align}
where $\bm{W}^{qkv}\in \mathbb{R}^{D\times3D_h}$, $\bm{b}^{qkv}\in \mathbb{R}^{3D_h}$, $\bm{W}^{o}\in \mathbb{R}^{h\cdot D_h\times D}$, $\bm{b}^{o}\in \mathbb{R}^{D}$, and $h$ is the number of the attention heads and $D_h$ is the feature size of each head.

The MLP module projects the features into a higher $D_f$-dimensional space to learn representations. Denoting the input to the MLP module as $\bm{Y}'\in \mathbb{R}^{N\times D}$, the calculation is as follows:
\begin{equation}
  \text{MLP}(\bm{Y}') = \text{GELU}(\bm{Y}' \bm{W}^1+\bm{b}^1)\bm{W}^2 + \bm{b}^2
\end{equation}
where $\bm{W}^1\in \mathbb{R}^{D\times D_f}$, $\bm{b}^1\in \mathbb{R}^{D_f}$, $\bm{W}^2\in \mathbb{R}^{D_f\times D}$, and $\bm{b}^2\in \mathbb{R}^{D}$.

As one can see, the large matrix multiplications contribute the most computational costs; hence, following previous works \cite{liu2021post,yuan2021ptq4vit}, we quantize all the weights and inputs of matrix multiplications, leaving LayerNorm and Softmax operations as floating-point types. Also, for efficient inference, we employ the hardware-friendly quantizers discussed below in the inference process.

\paragraph{Hardware-friendly quantizers} The uniform quantizer is one of the most popular choices that is well supported by the hardware, which is defined as:
\begin{align}
  Quant&: \bm{x}^{(\mathbb{Z})} = \text{clip}\left(\left\lfloor \frac{\bm{x}}{s} \right\rceil+z, 0, 2^b-1 \right) \\
  DeQuant&: \hat{\bm{x}} = s\left(\bm{x}^{(\mathbb{Z})}-z\right) \approx \bm{x}
\end{align}
where $\bm{x}$ and $\bm{x}^{(\mathbb{Z})}$ are the floating-point and quantized values, respectively, $\left\lfloor\cdot\right\rceil$ denotes the round function, and $b \in \mathbb{N}$ is the quantization bit-width. 
In the de-quantization procedure\footnote{In actual inference, the floating-point multiplication with $s$ is replaced by re-quantization to implement integer arithmetic.}, the de-quantized value $\hat{\bm{x}}$ approximately recovers $\bm{x}$.
Importantly, $s\in \mathbb{R}^+$ is the quantization scale and $z \in \mathbb{Z}$ is the zero-point, both of which are determined by the lower and upper bounds of $\bm{x}$ as follows:
\begin{equation}
\label{eq:sz}
s = \frac{\max(\bm{x})-\min(\bm{x})}{2^b-1}, \quad z = \left\lfloor-\frac{\min(\bm{x})}{s} \right\rceil
\end{equation}

The log2 quantizer is another common and hardware-oriented choice. Since it is only applied on post-Softmax activations in this paper, we just consider the quantization of positive values as follows:
\begin{align}
\label{eq:log_quant}
 Quant&: \bm{x}^{(\mathbb{Z})} = \text{clip}\left(\left\lfloor -\log_2 \frac{\bm{x}}{s} \right\rceil, 0, 2^b-1 \right) \\
\label{eq:log_dequant}
  DeQuant&: \hat{\bm{x}} = s\cdot 2^{-\bm{x}^{(\mathbb{Z})}} \approx \bm{x}
\end{align}
where both the log2 function and the base-2 power function can be implemented using the fast and efficient bit-shifting operations \cite{lee2017lognet,lin2021fq}.
   
For the application granularity of the above quantizers, channel-wise quantization for weights and layer-wise quantization for activations can balance accuracy and efficiency \cite{gholami2021survey,krishnamoorthi2018quantizing}, and are well supported by both hardware and software \cite{jacob2018quantization,yao2021hawq,li2022vit}, and thus have become a consensus in previous works \cite{yuan2021ptq4vit,ding2022towards}. In this paper, we follow the above quantization granularity in the inference process.

\begin{figure}[t]
    \centering
    \includegraphics[width=0.97\linewidth]{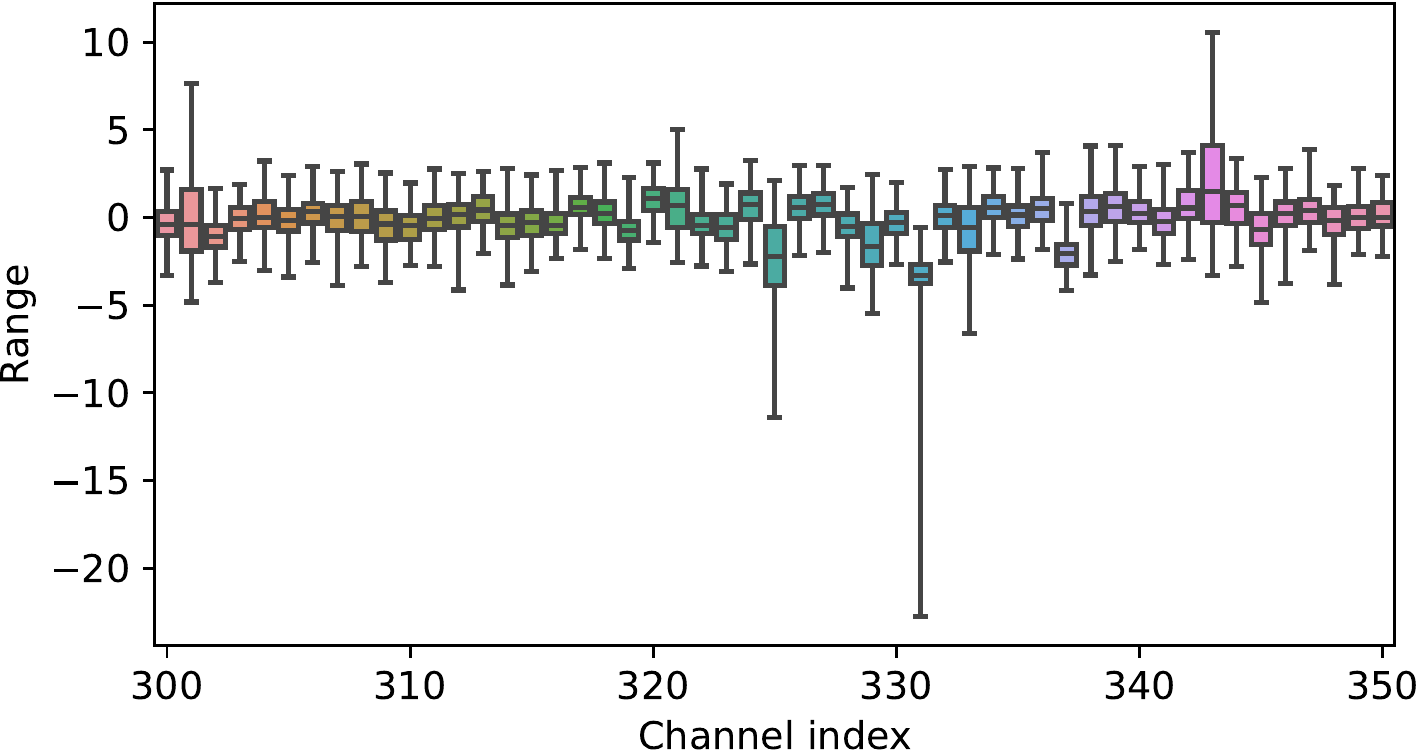}
    \caption{Boxplot of the 300th to 350th channels of the first module's post-LayerNorm activations in DeiT-S. As is evident, there is a severe inter-channel variation.}
\label{fig:LayerNorm}
\end{figure}

\subsection{Scale Reparam for LayerNorm Activations}
\label{sec:layernorm}

In ViTs, LayerNorm is applied  to normalize the input $\bm{X}\in \mathbb{R}^{N\times D}$ in the hidden feature dimension, and its calculation process is as follows:
\begin{equation}
\text{LayerNorm}(\bm{X}_{n,:}) = \frac{\bm{X}_{n,:}-\text{E}[\bm{X}_{n,:}]}{\sqrt{\text{Var}[\bm{X}_{n,:}]+\epsilon}}\odot \bm{\gamma} + \bm{\beta}
\end{equation}
where $n=1,2,\cdots,N$, $\text{E}[\bm{X}_{n,:}]$ and $\text{Var}[\bm{X}_{n,:}]$ are the mean and variance, respectively, and $\bm{\gamma}\in \mathbb{R}^D$ and $\bm{\beta}\in \mathbb{R}^D$ are the row vectors\footnote{In this paper, we make the convention that all vectors serve as row vectors by default to facilitate the formulation.} of linear affine factors. Here, $\odot$ denotes Hadamard product.

Looking into the post-LayerNorm activations, we find that they have a severe inter-channel variation, which is a critical limitation to the quantization performance.
More intuitively, the distribution boxplot of the 300th to 350th channels of the first module's post-LayerNorm activations in DeiT-S is illustrated in Figure \ref{fig:LayerNorm}, where the minimum, mean, and maximum ranges are 3.94, 7.11, and 22.2, respectively. In this case, layer-wise quantization that simply applys a unified quantization scale to each channel cannot accommodate such severe inter-channel variation, resulting in significant accuracy degradation. 
As an alternative, channel-wise quantization can address the above challenge.
However, channel-wise quantization for activations requires the support of the dedicated hardware and incurs additional computational overhead.

To address the above issues, we apply the quantization-inference decoupling paradigm and propose a scale reparameterization method for post-LayerNorm activations that transforms channel-wise quantization to layer-wise quantization, achieving both the accuracy of the former and the efficiency of the latter. Specifically, given the post-LayerNorm activations $\bm{X}'$, we first perform channel-wise quantization to obtain the quantization scale $\bm{s}\in R^{D}$ and zero-point $\bm{z}\in Z^{D}$.
Our goal is to reparameterize them to $\tilde{\bm{s}}=\tilde{s} \cdot \bm{1}$ and $\tilde{\bm{z}}=\tilde{z} \cdot \bm{1}$, where $\bm{1}$ is a $D$-dimensional row vector of all ones, and the scalars $\tilde{s}\in R^{1}$ and $\tilde{z}\in Z^{1}$ are ready for layer-wise quantization.
Here, $\tilde{s}$ and $\tilde{z}$ are pre-specified and we set them to the corresponding mean values in this paper, \ie, $\tilde{s}=\text{E}[\bm{s}], \tilde{z}=\text{E}[\bm{z}]$. Defining the variation factors $\bm{r}_1=\bm{s}/\tilde{\bm{s}}$ \footnote{In this paper, division between vectors is an element-wise operation like Hadamard product.} and $\bm{r}_2=\bm{z}-\tilde{\bm{z}}$, the following equations hold:
\begin{align}
  \label{eq:3.2-1} \tilde{\bm{z}} &= \bm{z}-\bm{r}_2 = \left\lfloor -\frac{\left[\min(\bm{X}'_{:,d})\right]_{1\leq d \leq D}+\bm{s}\odot \bm{r}_2}{\bm{s}} \right\rceil \\
  \label{eq:3.2-2} \tilde{\bm{s}} &= \frac{\bm{s}}{\bm{r}_1} = \frac{\left[\max(\bm{X}'_{:,d})-\min(\bm{X}'_{:,d})\right]_{1\leq d \leq D}/\bm{r}_1}{2^b-1}
\end{align}

Eq. \ref{eq:3.2-1} shows that adding $\bm{s}\odot \bm{r}_2$ to each channel of $\bm{X}'$ can yield $\tilde{\bm{z}}$, and Eq. \ref{eq:3.2-2} shows that dividing each channel of $\bm{X}'$ by $\bm{r}_1$ can yield $\tilde{\bm{s}}$. These operations can be achieved by adjusting the LayerNorm’s affine factors as follows:
\begin{equation}
    \label{eq:raparm_1}
  \widetilde{\bm{\beta}} = \frac{\bm{\beta}+\bm{s}\odot \bm{r}_2}{\bm{r}_1}, \quad \widetilde{\bm{\gamma}} = \frac{\bm{\gamma}}{\bm{r}_1}
\end{equation}

The above procedure accomplishes the reparameterization of $\tilde{\bm{s}}$ and $\tilde{\bm{z}}$, while this results in a distribution shift of activations, \ie, $\widetilde{\bm{X}}'_{n,:}=(\bm{X}'_{n,:}+\bm{s}\odot \bm{r}_2)/\bm{r}_1$. Fortunately, such distribution shift can be eliminated by the inverse compensation of the next layer's weights. To be specific, through equivalent transformations we have that:
\begin{equation}
\begin{split}
  \bm{X}'_{n,:}\bm{W}^{qkv}_{:,j}+\bm{b}^{qkv}_j = \frac{\bm{X}'_{n,:}+\bm{s}\odot \bm{r}_2}{\bm{r}_1} \left(\bm{r}_1\odot\bm{W}^{qkv}_{:,j}\right) \\ + \left(\bm{b}^{qkv}_j - (\bm{s}\odot \bm{r}_2) \bm{W}^{qkv}_{:,j}\right)
\end{split}
\end{equation}
Where $j=1,2,\cdots,3D_h$.
Thus, to align the next layer's outputs before and after the reparameterization, the weights can be adjusted as follows:
\begin{equation}
\label{eq:raparm_2}
\begin{aligned}
  \widetilde{\bm{W}}^{qkv}_{:,j} &= \bm{r}_1\odot\bm{W}^{qkv}_{:,j} \\
  \widetilde{\bm{b}}^{qkv}_j &= \bm{b}^{qkv}_j - (\bm{s}\odot \bm{r}_2) \bm{W}^{qkv}_{:,j}    
\end{aligned}
\end{equation}

Since the inter-channel variation factor $\bm{r}_1\in\mathbb{R}^D$ works in different dimensions from the quantization scale $\bm{s}^{qkv}\in\mathbb{R}^{3D_h}$ and zero-point $\bm{z}^{qkv}\in\mathbb{R}^{3D_h}$ of $\bm{W}^{qkv}$, the explicit solutions of the corresponding parameters of $\widetilde{\bm{W}}^{qkv}$ cannot be directly derived and need to be re-calibrated. It is worth noting that since the weights are inherently applied channel-wise quantization, the quantization performance is not sensitive to compensating $\bm{r}_1$ on the weights, making the re-calibration of $\widetilde{\bm{W}}^{qkv}$ incur only a slight accuracy loss.

In this way, by interpretable adjustment of the LayerNorm’s affine factors $\widetilde{\bm{\beta}}$ and $\widetilde{\bm{\gamma}}$ as well as the next layer’s weight $\widetilde{\bm{W}}^{qkv}$ and $\widetilde{\bm{b}}^{qkv}$, we confidently reparameterize the channel-wise quantization of $\bm{X}'$ with $\bm{s}$ and $\bm{z}$ to the layer-wise quantization with $\tilde{s}$ and $\tilde{z}$. And the adjustment strategy also works for the input $\bm{Y}'$ to the MLP module.
This easy-to-implement process allows us to fully benefit from the efficient inference of layer-wise quantization while featuring a robust characterization of the inter-channel variation that has only a slight performance drop compared to channel-wise quantization.

\begin{figure}[t]
    \centering
    \includegraphics[width=0.85\linewidth]{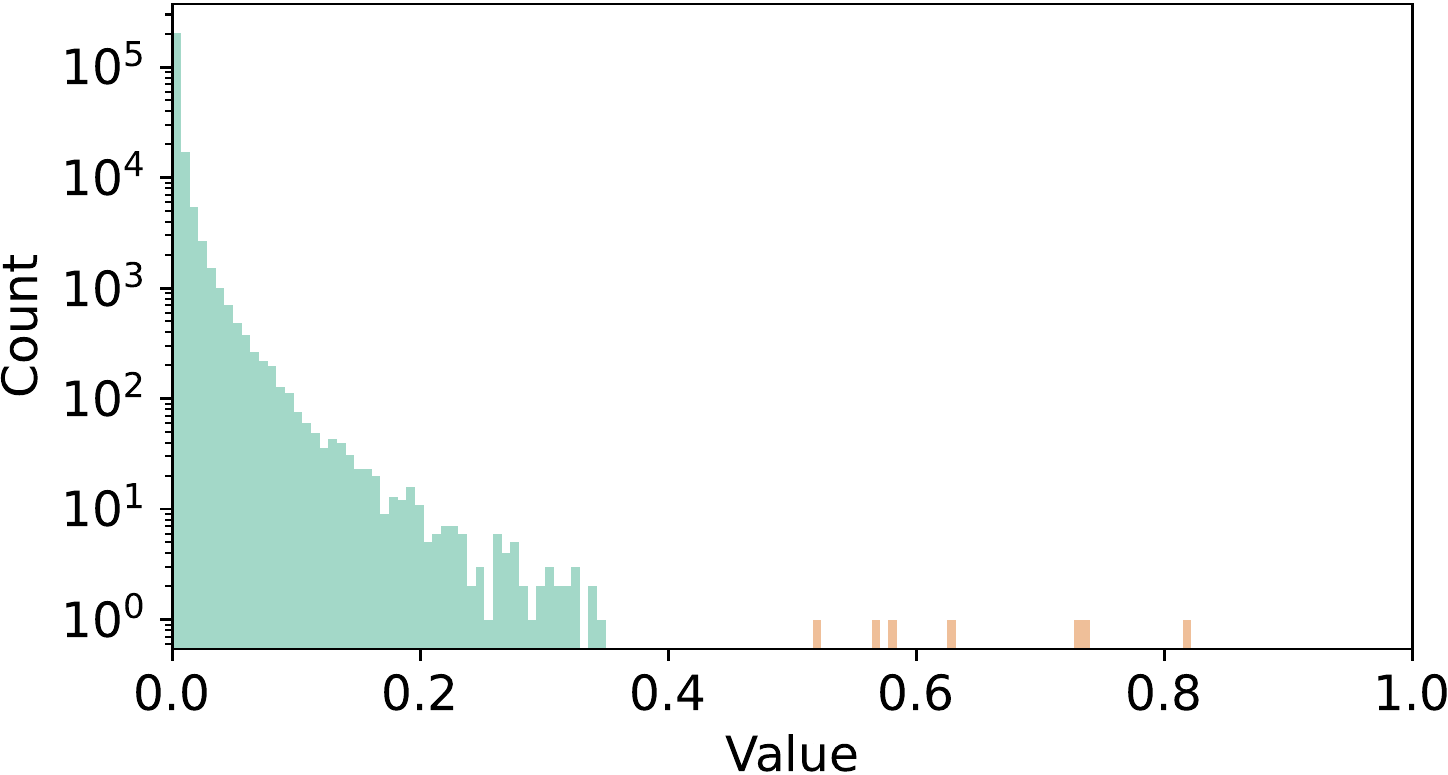}
    \caption{Histogram of the first MSA module's post-Softmax activations in DeiT-S. As one can see, it is extremely unbalanced, with the majority concentrated in small values (in green) and a few scattered in large values (in orange).}
\label{fig:Softmax}
\end{figure}

\subsection{Scale Reparam for Softmax Activations}
\label{sec:softmax}
In ViTs, the Softmax operation converts the attention scores of the MSA module into probabilities, bounding the values to the (0, 1) interval.
However, these probabilities, termed as post-Softmax activations, have a power-law distribution far from the Gaussian and Laplace distributions, which is extremely unbalanced and thus is identified as another key obstacle to quantization. For instance, Figure \ref{fig:Softmax} shows the distribution histogram of the first MSA module's post-Softmax activations in DeiT-S. It can be observed that the majority of activations are concentrated in relatively small values, and only a few activations are discretely scattered in large values (close to 1). Statistically, even 99.2\% of the activations are smaller than 0.3. It should be noted that the remaining 0.8\% of activations cannot be viewed as outliers to be naively clipped; instead, these values reflect important correlations between patches that guide the MSA module to give more attention, thus we have to preserve them well in the quantization process.

To deal with the above power-law distribution, previous work \cite{lin2021fq} directly applies the log2 quantizer depending on hardware efficiency considerations. Despite the better performance than the uniform quantizer, the log2 quantizer still fails to provide a reliable and robust description of the distribution in practice. Taking the simple case of $s=1$ as an example, the log2 quantizer takes values at levels \{$2^0$, $2^{-1}$, $2^{-2}$, $\cdots$\}, and according to Eq. \ref{eq:log_quant}, the values in the relatively large interval [$2^{-1.5}$, $2^{-0.5}$], \ie, [0.354, 0.707], are in principle rounded to $2^{-1}$. This overly sparse description of important attention scores greatly weakens the representational power of the MSA module. In contrast, the log$\sqrt{2}$ quantizer, which provides a higher quantization resolution for large values, can describe the distribution in a more accurate fashion.
Nevertheless, it is unfriendly to hardware and fails to benefit from efficient bit-shifting operations in inference as the log2 quantizer does.

Inspired by the quantization-inference decoupling paradigm, we are motivated to explore how to convert the log$\sqrt{2}$ quantizer to log2 quantizer. With it, we can enjoy both the high accuracy of the former and the bit-shifting operations of the latter. To this end, the base changing methods are designed for the quantization and de-quantization procedures, respectively.
First, given the post-Softmax activations $\bm{A}$ and the log$\sqrt{2}$ quantizer's scale $s\in \mathbb{R}^1$, according to the base changing formula of the log function we have:
\begin{equation}
\label{eq:3.3-1}
\begin{split}
    \bm{A}^{(\mathbb{Z})} &= \text{clip}\left(\left\lfloor -\log_{\sqrt{2}} \frac{\bm{A}}{s} \right\rceil, 0, 2^b-1 \right) \\ 
    &= \text{clip}\left(\left\lfloor -2\log_2 \frac{\bm{A}}{s} \right\rceil , 0, 2^b-1 \right)
\end{split}
\end{equation}

Thus, for the quantization procedure, the conversion to the log2 quantizer can be achieved by simply multiplying by a constant factor.
Similarly, in the de-quantization procedure, the base changing formula of the pow function is utilized to obtain the base-2 form; however, the new exponential term $-\frac{\bm{A}^{(\mathbb{Z})}}{2}$ is not guaranteed to be an integer that is necessary to perform the bit-shifting operations. Therefore, we discuss the parity of $-\frac{\bm{A}^{(\mathbb{Z})}}{2}$ by case as follows:
\begin{equation}
\begin{split}
    \widehat{\bm{A}} &= s\cdot \sqrt{2}^{-\bm{A}^{(\mathbb{Z})}} = s\cdot 2^{-\frac{\bm{A}^{(\mathbb{Z})}}{2}} \\
    &= \begin{cases}
        s\cdot 2^{-\frac{\bm{A}^{(\mathbb{Z})}}{2}} &\;\; \bm{A}^{(\mathbb{Z})}=2k, k\in \mathbb{Z} \\
        s\cdot 2^{-\frac{\bm{A}^{(\mathbb{Z})+1}}{2}}\cdot \sqrt{2} &\;\; \bm{A}^{(\mathbb{Z})}=2k+1, k\in \mathbb{Z}
    \end{cases} \\
    &= s\cdot 2^{\left\lfloor-\frac{\bm{A}^{(\mathbb{Z})}}{2}\right\rfloor}
    \cdot \left[\mathds{1}(\bm{A}^{(\mathbb{Z})})\cdot(\sqrt{2}-1)+1 \right]
\end{split}
\end{equation}
where $\left\lfloor\cdot\right\rfloor$ denotes the floor function, $\left\lfloor-\frac{\bm{A}^{(\mathbb{Z})}}{2}\right\rfloor$ is consistently an integer, and $\mathds{1}(\cdot)$ is a parity indicator function that is 0 at even numbers and 1 at odd numbers.

The above parity indicator function and its coefficients can be merged into $s$ to obtain the reparameterized scale $\tilde{s}$ as follows:
\begin{equation}
\label{eq:3.3-2}
    \tilde{s} = s\cdot \left[\mathds{1}(\bm{A}^{(\mathbb{Z})})\cdot(\sqrt{2}-1)+1 \right]
\end{equation}

Eventually, thanks to the reparameterization of $\tilde{s}$, the de-quantization procedure is also able to benefit from the efficient bit-shifting operations. Note that compared to the previous scale $s$, the reparameterized scale $\tilde{s}$ only introduces a slight additional computational overhead in the inference process, due to the fact that the parity indicator function can be computed with great efficiency, \eg, by simply querying the least significant bit of $\bm{A}^{(\mathbb{Z})}$ on FPGAs. 

\section{Experiments}
\subsection{Experimental Setup}
\paragraph{Models and datasets}
For the image classification task, RepQ-ViT is evaluated on ImageNet \cite{krizhevsky2012imagenet} dataset with different model variants: ViT \cite{dosovitskiy2020image}, DeiT \cite{touvron2021training}, and Swin \cite{liu2021swin}. 
For the object detection and instance segmentation tasks, RepQ-ViT is evaluated on COCO \cite{lin2014microsoft} dataset using two typical frameworks: Mask R-CNN \cite{he2017mask} and Cascade Mask R-CNN \cite{cai2018cascade} with Swin \cite{liu2021swin} as the backbone.

\paragraph{Implementation details} 
All pretrained full-precision models are obtained from Timm\footnote{https://github.com/rwightman/pytorch-image-models} library. For a fair comparison with the previous works \cite{yuan2021ptq4vit,ding2022towards}, we randomly select 32 samples from ImageNet dataset for image classification and 1 sample from COCO dataset for object detection and instance segmentation to calibrate the quantization parameters. For the calibration strategy, we apply the prevalent Percentile \cite{li2019fully} method, with channel-wise quantization for weights and layer-wise quantization for activations in the inference process. Scale reparameterization is applied to post-LayerNorm activations in all blocks (including those in PatchMerging layers of Swin) and to post-Softmax activations in all MSA modules.
Note that our proposed RepQ-ViT is free of any hyperparameters and thus offers a high ease of implementation and generality, which is significantly superior to existing methods.

\subsection{Quantization Results on ImageNet Dataset}

\begin{table*}[t]
\centering
\small
\begin{tabular}{ccccccccccc}
\toprule
\textbf{Method} & \textbf{No HP} & \textbf{No REC} &\textbf{Prec. (W/A)} & \textbf{ViT-S }& \textbf{ViT-B} & \textbf{DeiT-T }& \textbf{DeiT-S} & \textbf{DeiT-B} & \textbf{Swin-S }& \textbf{Swin-B} \\
\midrule
Full-Precision & - & - & 32/32 & 81.39 & 84.54 & 72.21 & 79.85 & 81.80 & 83.23 & 85.27 \\
\midrule
FQ-ViT \cite{lin2021fq} & $\times$ & $\checkmark$ & 4/4 & 0.10 & 0.10 & 0.10 & 0.10 & 0.10 & 0.10 & 0.10 \\
PTQ4ViT \cite{yuan2021ptq4vit} & $\times$ & $\times$ & 4/4 & 42.57 & 30.69 & 36.96 & 34.08 & 64.39 & 76.09 & 74.02 \\
APQ-ViT \cite{ding2022towards} & $\times$ & $\times$ & 4/4 & 47.95 & 41.41 & 47.94 & 43.55 & 67.48 & 77.15 & 76.48 \\
RepQ-ViT (ours) & $\checkmark$ & $\checkmark$ & 4/4 & \textbf{65.05} & \textbf{68.48} &\textbf{57.43} & \textbf{69.03} & \textbf{75.61} & \textbf{79.45} & \textbf{78.32} \\
\midrule
FQ-ViT \cite{lin2021fq} & $\times$ & $\checkmark$ & 6/6 & 4.26 & 0.10 & 58.66 & 45.51 & 64.63 & 66.50 & 52.09 \\
PSAQ-ViT \cite{li2022patch} & $\times$ & $\checkmark$ & 6/6 & 37.19 & 41.52 & 57.58 & 63.61 & 67.95 & 72.86 & 76.44 \\
Ranking \cite{liu2021post} & $\times$ & $\times$ & 6/6 & - & 75.26 & - & 74.58 & 77.02 & - & - \\
PTQ4ViT \cite{yuan2021ptq4vit} & $\times$ & $\times$ & 6/6 & 78.63 & 81.65 & 69.68 & 76.28 & 80.25 & 82.38 & 84.01 \\
APQ-ViT \cite{ding2022towards} & $\times$ & $\times$ & 6/6 & 79.10 & 82.21 & 70.49 & 77.76 & 80.42 & 82.67 & 84.18 \\
RepQ-ViT (ours) & $\checkmark$ & $\checkmark$ & 6/6 & \textbf{80.43} & \textbf{83.62} & \textbf{70.76} & \textbf{78.90} & \textbf{81.27} & \textbf{82.79} & \textbf{84.57} \\
\bottomrule
\end{tabular}
\vspace{2pt}
\caption{Quantization results of image classification on ImageNet dataset, where each data presents the Top-1 accuracy (\%) obtained by quantizing each model. Here, we abbreviate ``No Hyperparameters'' as ``No HP" and ``No Reconstruction'' as ``No REC", and ``Prec. (W/A)'' indicates that the quantization bit-precision of the weights and activations are W and A bits, respectively.\\}
\label{tab:imagenet}
\end{table*}

\begin{table*}[t]
\centering
\small
\begin{tabular}{cccccccccccc}
\toprule
\multirow{3.5}{*}{\textbf{Method}} & \multirow{3.5}{*}{\textbf{No HP}} & \multirow{3.5}{*}{\textbf{No REC}} & \multirow{3.5}{*}{\textbf{Prec. (W/A)}} & \multicolumn{4}{c}{\textbf{Mask R-CNN}} & \multicolumn{4}{c}{\textbf{Cascade Mask R-CNN}} \\
\cmidrule(lr){5-8}\cmidrule(lr){9-12}
&&&& \multicolumn{2}{c}{\textbf{w. Swin-T}} & \multicolumn{2}{c}{\textbf{w. Swin-S}} & \multicolumn{2}{c}{\textbf{w. Swin-T}} & \multicolumn{2}{c}{\textbf{w. Swin-S}} \\
&&&& AP$^\text{box}$ & AP$^\text{mask}$ & AP$^\text{box}$ & AP$^\text{mask}$ & AP$^\text{box}$ & AP$^\text{mask}$ & AP$^\text{box}$ & AP$^\text{mask}$ \\
\midrule
Full-Precision & - & - & 32/32 & 46.0 & 41.6 & 48.5 & 43.3 & 50.4 & 43.7 & 51.9 & 45.0 \\
\midrule
PTQ4ViT \cite{yuan2021ptq4vit} & $\times$ & $\times$ & 4/4 & 6.9 & 7.0 & 26.7 & 26.6 & 14.7 & 13.5 & 0.5 & 0.5  \\
APQ-ViT \cite{ding2022towards} & $\times$ & $\times$ & 4/4 & 23.7 & 22.6 & \textbf{44.7} & 40.1 & 27.2 & 24.4 & 47.7 & 41.1 \\
RepQ-ViT (ours) & $\checkmark$ & $\checkmark$ & 4/4 & \textbf{36.1} & \textbf{36.0} & 44.2 & \textbf{40.2}& \textbf{47.0} & \textbf{41.4} & \textbf{49.3} & \textbf{43.1} \\
\midrule
PTQ4ViT \cite{yuan2021ptq4vit} & $\times$ & $\times$ & 6/6 & 5.8 & 6.8 & 6.5 & 6.6 & 14.7 & 13.6 & 12.5 & 10.8 \\
APQ-ViT \cite{ding2022towards} & $\times$ & $\times$ & 6/6 & \textbf{45.4} & 41.2 & \textbf{47.9} & 42.9 & 48.6 & 42.5 & 50.5 & 43.9 \\
RepQ-ViT (ours) & $\checkmark$ & $\checkmark$ & 6/6 & 45.1 & \textbf{41.2} & 47.8 & \textbf{43.0} & \textbf{50.0} & \textbf{43.5} & \textbf{51.4} & \textbf{44.6} \\
\bottomrule

\end{tabular}
\vspace{1.5pt}
\caption{Quantization results of object detection and instance segmentation on COCO dataset. Here, ``AP$^\text{box}$'' is the box average precision for object detection, and ``AP$^\text{mask}$'' is the mask average precision for instance segmentation.}
\label{tab:coco}
\end{table*}

We start by comparing the quantization results of the proposed RepQ-ViT and existing methods on ImageNet dataset for image classification, as reported in Table \ref{tab:imagenet}. 
It is worth noting that hyperparameters and reconstruction procedures are also explicitly presented in the Table as conditional indicators. Thanks to the non-dependence on these two indicators, RepQ-ViT is believed to be more practical and general in real-world applications. We focus on the performance of low-bit quantization, including W4/A4 and W6/A6 quantization, to highlight the advantages of RepQ-ViT. In W4/A4 quantization, previous works all suffer from non-trivial performance degradation. For instance, FQ-ViT becomes infeasible with only 0.1\% accuracy, while PTQ4ViT and APQ-ViT improve accuracy with the help of reconstruction but remain far from practical usability. Fortunately, RepQ-ViT can maintain the data distribution through the design of complex quantizers to achieve robust and substantial improvement of the quantization performance, with encouraging 27.07\% and 25.48\% improvement over APQ-ViT in ViT-B and DeiT-S quantization, respectively. When quantizing DeiT-B, Swin-S, and Swin-B, RepQ-ViT consistently obtains an interesting decrease in accuracy of less than 7\%. To the best of our knowledge, we are the first to break the limit of 4-bit PTQ of ViTs to the usable level. 
In addition, in W6/A6 quantization, RepQ-ViT can achieve an accuracy comparable to that of the full-precision baseline with a model size compressed by 5.3 times. In DeiT-B and Swin-S quantization, RepQ-ViT achieves 81.27\% and 82.79\% accuracy, respectively, with only 0.53\% and 0.44\% accuracy loss.

\subsection{Quantization Results on COCO Dataset}
The object detection and instance segmentation experiments are conducted on COCO dataset, and the quantization results are shown in Table \ref{tab:coco}. As before, we also explicitly list whether hyperparameters and reconstruction procedures are required. 
Due to the more complex model architectures in high-level tasks, PTQ4ViT's twin-scale search loses its viability, leading to disappointing quantization performance. APQ-ViT is not robust to different backbones; it yields good results with Swin-S as the backbone while it causes severe performance degradation when Swin-T serves as the backbone, for instance, in W4/A4 quantization of Cascade Mask R-CNN framework with Swin-T, box AP and mask AP are degraded by 23.2 and 19.3, respectively. This greatly limits the practical deployment and application of the quantized models. 
Compared with previous methods, our proposed RepQ-ViT achieves more advanced performance with high robustness. When performing W4/A4 quantization in the case of Swin-T backbone, for Mask R-CNN framework, RepQ-ViT improves over APQ-ViT by 12.4 box AP and 13.4 mask AP; for Cascade Mask R-CNN framework, RepQ-ViT yields a boost of 19.8 box AP and 17.0 mask AP over APQ-ViT.
Moreover, in W6/A6 quantization, RepQ-ViT produces only a slight accuracy loss over the full-precision baseline. When quantizing Cascade Mask R-CNN framework with Swin-T, RepQ-ViT reached 50.0 box AP and 43.5 mask AP, which is just 0.4 box AP and 0.2 mask AP lower than the full-precision baseline. Similar results can also be obtained when Swin-S serves as the backbone, achieving 51.4 box AP and 44.6 mask AP.

\subsection{Ablation Studies}
To validate the effectiveness of the main components of the proposed RepQ-ViT framework, we perform two ablation studies of the scale reparameterization methods for post-LayerNorm and post-Softmax activations, respectively, as shown in Tables \ref{tab:ablation_1} and \ref{tab:ablation_2}.

\begin{table}[t]
\centering
\small
\begin{tabular}{cccc}
\toprule
\textbf{Model} & \textbf{Method} & \textbf{Hardware} &\textbf{Top-1 (\%)} \\
\midrule
\multirow{4.5}{*}{DeiT-S} & Full-Precision & - & 79.85 \\
\cmidrule{2-4}
& Layer-Wise Quant. & $\checkmark$ & 33.17 \\
& Channel-Wise Quant. & $\times$ & \textbf{70.28} \\
& Scale Reparam (ours) & $\checkmark$ & 69.03 \\
\midrule
\multirow{4.5}{*}{Swin-S} & Full-Precision & - & 83.23 \\
\cmidrule{2-4}
& Layer-Wise Quant. & $\checkmark$ & 57.63 \\
& Channel-Wise Quant. & $\times$ & \textbf{80.52} \\
& Scale Reparam (ours) & $\checkmark$ & 79.45 \\
\bottomrule

\end{tabular}
\vspace{1.5pt}
\caption{Ablation studies of different quantizers (W4/A4) for post-LayerNorm activations. Here, ``Hardware'' indicates whether the obtained quantized model is hardware-friendly and can be efficiently computed in inference.}
\label{tab:ablation_1}
\end{table}

\begin{table}[t]
\centering
\small
\begin{tabular}{cccc}
\toprule
\textbf{Model} & \textbf{Method} & \textbf{Hardware} &\textbf{Top-1 (\%)} \\
\midrule
\multirow{4.5}{*}{DeiT-S} & Full-Precision & - & 79.85 \\
\cmidrule{2-4}
& Log2 Quant. & $\checkmark$ & 67.71 \\
& Log$\sqrt{2}$ Quant. & $\times$ & \textbf{69.03} \\
& Scale Reparam (ours) & $\checkmark$ & \textbf{69.03} \\
\midrule
\multirow{4.5}{*}{Swin-S} & Full-Precision & - & 83.23 \\
\cmidrule{2-4}
& Log2 Quant. & $\checkmark$ & 77.87 \\
& Log$\sqrt{2}$ Quant. & $\times$ & \textbf{79.45} \\
& Scale Reparam (ours) & $\checkmark$ & \textbf{79.45} \\
\bottomrule

\end{tabular}
\vspace{1.5pt}
\caption{Ablation studies of different quantizers (W4/A4) for post-Softmax activations.}
\label{tab:ablation_2}
\end{table}

Table \ref{tab:ablation_1} reports the ablation study results of different quantizers (W4/A4) for post-LayerNorm activations. Taking DeiT-S as an example, direct layer-wise quantization cannot represent the data distribution well and achieves only 33.17\% accuracy. Applying channel-wise quantization can solve the above issue with 70.28\% accuracy; however, it fails to satisfy the hardware characteristics to enable efficient calculations in the inference process. Therefore, the scale reparameterization method, which converts channel-wise quantization to layer-wise quantization, can allow for both high accuracy and efficient inference. Note that due to the recalibration of $\widetilde{\bm{W}}^{qkv}$, the scale reparameterization method produces a slight performance drop (1.25\%) compared to channel-wise quantization.

The results of different quantizers (W4/A4) for post-Softmax activations are reported in Table \ref{tab:ablation_2}. Log$\sqrt{2}$ quantizers can better fit the extreme distributions of attention scores, and in particular, has better a representation of scattered large values, thus providing a 1.58\% improvement in accuracy over simple log2 quantizers in the case of Swin-S quantization. To solve the inefficiency problem of log$\sqrt{2}$ quantizers, scale reparameterization is applied to accomplish the conversion to log2 quantizers. Here, the scale reparameterization method for post-Softmax activations employs exactly equivalent transformations and thus yields the same accuracy as log$\sqrt{2}$ quantizers, at the cost of only a slight additional computational overhead in inference compared to log2 quantizers.

\subsection{Efficiency Analysis}
We also compare the efficiency of different methods, including the data quantity and time consumption required for quantization calibration, as shown in Table \ref{tab:ablation_3}. Here, time consumption is measured on a single 3090 GPU. Since there is no reconstruction in FQ-ViT, the quantized models can be obtained rapidly, however, the performance drops severely even with 1000 samples for calibration.
Our RepQ-ViT requires only 32 samples as PTQ4ViT, while it is free of reconstruction and thus can yield quantized models with higher accuracy more quickly compared to PTQ4ViT.

\begin{table}[t]
\centering
\tabcolsep=1mm
\small
\begin{tabular}{ccccc}
\toprule
\textbf{Model} & \textbf{Method} & \textbf{Top-1 (\%)} & \textbf{Calib Data} & \textbf{GPU Min.} \\
\midrule
\multirow{4.5}{*}{DeiT-S} & Full-Precision & 79.85 & - & - \\
\cmidrule{2-5}
& FQ-ViT \cite{lin2021fq} & 0.10 & 1000 & \textbf{0.5} \\
& PTQ4ViT \cite{yuan2021ptq4vit} & 34.08 & \textbf{32} & 3.2 \\
& RepQ-ViT (ours) & \textbf{69.03} & \textbf{32} & 1.3 \\
\midrule
\multirow{4.5}{*}{Swin-S} & Full-Precision & 83.23 & - & -  \\
\cmidrule{2-5}
& FQ-ViT \cite{lin2021fq} & 0.10 & 1000 & \textbf{1.1} \\
& PTQ4ViT \cite{yuan2021ptq4vit} & 76.09 & \textbf{32} & 7.7 \\
& RepQ-ViT (ours) & \textbf{79.45} & \textbf{32} & 2.9 \\
\bottomrule

\end{tabular}
\caption{Comparison of the data quantity and time consumption (in minutes) during the quantization (W4/A4) calibration.}
\label{tab:ablation_3}
\end{table}

\section{Conclusions}
In this paper, we propose RepQ-ViT, a novel post-training quantization framework for vision transformers. RepQ-ViT applies the quantization-inference decoupling paradigm, where complex quantizers are employed in the quantization process and simple hardware-friendly quantizers are employed in the inference process, and both are explicitly bridged by scale reparameterization. More specifically, RepQ-ViT resolves the extreme distributions of two components: for post-LayerNorm activations with severe inter-channel variation, channel-wise quantization is initially applied and then is reparameterized to layer-wise quantization; for post-Softmax activations with power-law features, log$\sqrt{2}$ quantization is initially applied and then is reparameterized to
log2 quantization. Exhaustive experiments are performed to fully validate the superiority of RepQ-ViT, showing that it significantly outperforms existing methods in low-bit quantization.

In the future, one can extend the reparameterization of channel-wise to layer-wise quantization to more activations. One can also try to combine log$\sqrt{2}$ and log2 quantization to better describe the power-law distribution.

\section*{Acknowledgement}
This work was supported in part by the National Key Research and Development Program of China under Grant 2022ZD0119402; in part by the National Natural Science Foundation of China under Grant 62276255.

{\small
\bibliographystyle{ieee_fullname}
\bibliography{egbib}
}

\end{document}